\documentclass[runningheads]{llncs}

\usepackage{eccv}

\usepackage{eccvabbrv}

\usepackage{graphicx}
\usepackage{booktabs}

\usepackage[accsupp]{axessibility}  

\usepackage{hyperref}

\usepackage{orcidlink}
\usepackage[dvipsnames]{xcolor}
\usepackage{graphicx}
\usepackage{algorithm}
\usepackage{algpseudocode}
\usepackage{standalone}
\usepackage{algorithm}
\usepackage{algpseudocode}

\usepackage{subcaption}
\DeclareCaptionSubType*{algorithm}

\DeclareCaptionLabelFormat{alglabel}{Alg.~#2}

\newcommand{\C}{\mathbb{C}}
\newcommand{\R}{\mathbb{R}}
\newcommand{\E}{\mathbb{E}}

\newcommand{\real}{\mathcal{R}}
\newcommand{\imag}{\mathcal{I}}

\usepackage{tikz}
\usetikzlibrary{calc, positioning, patterns, arrows.meta}
\newcommand*{\TickSize}{0.1}

\DeclareMathOperator{\argmax}{argmax}
\DeclareMathOperator{\argmin}{argmin}
\DeclareMathOperator{\sort}{sort}

\begin{document}

\title{DeepCSHAP: Utilizing Shapley Values to Explain Deep Complex-Valued Neural Networks} 

\titlerunning{DeepCSHAP: Utilizing Shapley Values to Explain CVNNs}

\author{Florian Eilers\inst{1}\orcidlink{0000-0003-0726-3287} \and
Xiaoyi Jiang \inst{1}\orcidlink{0000-0001-7678-9528}}

\authorrunning{F. Eilers and X. Jiang}

\institute{Department of Computer Science, University of Münster, Germany
\email{\{florian.eilers,xjiang\}@uni-heidelberg.de}}

\maketitle

\begin{abstract}
Deep Neural Networks are widely used in academy as well as corporate and public applications, including safety critical applications such as health care and autonomous driving.
The ability to explain their output is critical for safety reasons as well as acceptance among applicants. A multitude of methods have been proposed to explain real-valued neural networks. 
Recently, complex-valued neural networks have emerged as a new class of neural networks dealing with complex-valued input data without the necessity of projecting them onto $\R^2$.
This brings up the need to develop explanation algorithms for this kind of neural networks.
In this paper we provide these developments.
While we focus on adapting the widely used DeepSHAP algorithm to the complex domain, we also present versions of four gradient based explanation methods suitable for use in complex-valued neural networks.
We evaluate the explanation quality of all presented algorithms and provide all of them as an open source library adaptable to most recent complex-valued neural network architectures.
\end{abstract}    

\section{Introduction}
\label{sec:intro}

Many applications profit from the success of deep neural networks. 
Most applications deal with real-valued numbers, such as natural images, encodings for natural language or real-valued signals. The standard real-valued deep neural networks are suitable to solve problems with these kinds of data.
There is a class of applications, however, that naturally deal with complex numbers in the data, such as MRI \cite{cole2021analysis}, PolSAR satellite images \cite{barrachina2023impact} or Fourier transforms of real-valued signals for frequency analysis \cite{Trabels2018deep}.
These kinds of data motivate the use of complex-valued neural networks (CVNNs), where all building blocks of deep neural networks are translated to the complex domain to directly process these data without use of an isomorphism from $\C$ to $\R^2$. 
It has been shown that in many applications (but not all \cite{yin2020phasen}) complex-valued neural networks show superior performance, especially when dealing with complex-valued input data but sometimes even when dealing with real-valued input data \cite{ko2022coshnet}.

On major drawback of real- and complex-valued deep learning architectures is their lack of interpretation. Their (often) millions of parameters make them hard to understand and interpret, leading to the often referred to black box.
To (partially) solve this issue, many methods in the regime of explainable artificial intelligence (XAI) have been proposed for real-valued neural networks.
Early works in this regime work with explanations based on gradients \cite{simonyan2013deep, springenberg2014striving, sundararajan2016gradients} while more recent works make use of more sophisticated methods, such as local linearization of the model \cite{ribeiro2016should, shrikumar2017learning} or Shapley values - a measure for cooperative games from game theory \cite{vstrumbelj2014explaining}.

To the best of our knowledge, no explanation methods have been proposed to explain complex-valued neural networks. 
Directly applying real-valued explanation algorithms to complex-valued architectures is often impossible and the adaptions are non trivial.
In this paper, we aim to fill this gap by developing a complex-valued variant of the widely used DeepSHAP \cite{lundberg2017unified} algorithm.
We provide proofs for the complex-valued chain rule used within the algorithm as well as for some desirable properties that reduce computational complexity.
Additionally, we adapt four gradient based explanation methods to the complex domain.
We then evaluate explanation quality of all of these methods on two datasets - one real-valued and one complex-valued dataset.
Additionally, we validate our theoretical findings on these two datasets. 

The key contributions of our work are:
\begin{enumerate}
    \item Development of a complex-valued DeepSHAP algorithm, including the proof of a complex-valued chain rule.
    \item Adaptation of four gradient based explanation methods to the complex domain. 
    \item Evaluation of explanation quality of these five methods on two complex-valued neural networks applied to a real-valued and a complex-valued dataset.
    \item Validation of theoretical results on these two datasets.
\end{enumerate}
The source code with a pytorch library containing all the explanation methods in \autoref{sec:cv_grad} and \autoref{sec:deepcshap}, as well as all experiments from \autoref{sec:eval} is publicly available\footnote{In the supplementary material for reviewers, online after publication.}.

\section{Related work} \label{sec:rel_work}
\subsection{Explanation methods}
Early explanation methods are mostly based on gradients. Simonyan et al. \cite{simonyan2013deep} use gradients as an explanation method. This idea is then extended to Guided Backpropagation \cite{springenberg2014striving}, where negative gradients are zeroed out throughout the backward pass.
Sundararajan et al. \cite{sundararajan2016gradients} propose the integrated gradients methods, where instead of using the plain gradients, they integrate the gradients over an interpolation from a starting value (e.g. an all zeros image) to the actual image.
Lastly, Bach et al. propose Layerwise Relevance Propagation \cite{bach2015pixel}, which is not a gradient based method per se, but is equivalent to computing the gradient times the input under certain assumptions \cite{kindermans2016investigating}.

LIME by Riberiro et al. \cite{ribeiro2016should} aims to locally explain a model's output by fitting a linear model in a neighborhood of a certain input. This local linear model is easily explainable and is used to locally explain the more complex model.

DeepLIFT by Shrikumar et al. \cite{shrikumar2017learning} introduces an explanation method that only needs one backward pass to explain a model output for a specific input.
They define heuristic rules how to explain single layers and then define a chain rule on how to explain a full network as a concatenation of these layers. 

SHAP \cite{lundberg2017unified} is a framework that defines a class of explanation methods - additive feature attribution methods - by an additive property, which is already satisfied by LIME and DeepLIFT. They formulate three axioms based on Shapley values from game theory and show that their defined SHAP values is the only explanation method to fulfill all of those axioms. They then derive kernelSHAP, a combination of Linear LIME and Shapley values as well as DeepSHAP, which makes use of the chain rule from DeepLIFT but replaces the heuristic rules with SHAP values for every layer. A complex-valued adaptation of DeepSHAP is the main focus of this work, so DeepSHAP will be explained in detail in \autoref{subsec:deepshap}.
Many variants of these methods have been proposed and are widely used in the literature \cite{adadi2018peeking}.

\subsection{Complex-valued neural networks}
After early works by Hirose et al. \cite{hirose2003complex}, the main building blocks for complex-valued convolutional neural networks have been explored by Trabelsi et al. \cite{Trabels2018deep}.
More recently there have also been developments of a complex-valued transformer architecture \cite{yang2020complex, eilers2023building} and many more architectures \cite{lee2022complex}.
In the last two years, besides numerous applications, there have been multiple theoretical works on complex-valued neural networks, such as analysis of training behavior \cite{tan2022real} and the ability to generalize \cite{chen2023spectral} as well as hardware optimizations for complex-valued calculations \cite{zhang2021optical}.
A multitude of complex-valued architectures have been used for applications such as MRI reconstruction \cite{cole2021analysis, vasudeva2022compressed}, PolSAR image processing \cite{li2020sscv, barrachina2023impact}, speech enhancement \cite{choi2018phase, hu2020dccrn}. A broad overview over the field can be found in \cite{bassey2021survey, lee2022complex}.

To the best of our knowledge, there exist no works on explainability complex-valued neural networks. Mayer et al. \cite{mayer2022deep} mention that they are not able to explain their model, since SHAP in its current form is not suitable for complex-valued neural networks. 
There are works that use explainability methods for real-valued neural networks to process complex-valued data \cite{lin2022sspnet}, which is a different approach that does not benefit from the success of complex-valued neural networks and is thus out of scope of this paper.
\section{Real-valued explaination methods}\label{sec:rv_xai}
\subsection{Gradient based methods}
\label{subsec:rv_grad}
A straightforward method for model explanation is to use the gradient of the output w.r.t to the input pixels as an explanation as proposed in \cite{simonyan2013deep}.
This can be modified to integrated gradients, where the input image is interpolated from some starting value (e.g. a black/all zeros image) and the gradients of all interpolation steps are integrated \cite{sundararajan2016gradients}. 
Additionally, by multiplying the gradients with the input image an approximation of Layerwise Relevance Propagation \cite{bach2015pixel} can be computed, as shown by Kindermans et al. \cite{kindermans2016investigating}.

Lastly Guided Backpropagation \cite{springenberg2014striving} uses gradients as an explanation but zeros out negative gradients (i.e. applies ReLU on the gradients) throughout the backward pass.

\subsection{DeepSHAP}\label{subsec:deepshap}
Lundberg et al. \cite{lundberg2017unified} define a class of explanation methods - additive feature attribution methods - as explanation models that fulfill \autoref{eq:ad_feat}. 
For some fixed input $x$, let $f_x$ be a (local) explanation model, which is a simplified version of the original model that acts on a simplified binary model input defined by a local mapping $ h_x(z') = z$ while assuring $f_x(z') \approx f(h_x(z'))$ if $z' \approx x'$.
An additive feature attribution method is then defined as a model fulfilling the equation:
\begin{equation}\label{eq:ad_feat}
    f_x(z') = \phi_0 + \sum_{j=1}^n \phi_j z'_j
\end{equation}
Here $z'$ is a binary variable describing the presence or absence of features of the simplified input.
They show that multiple methods from the literature \cite{ribeiro2016should, bach2015pixel, shrikumar2017learning, vstrumbelj2014explaining} already fulfill this definition.
They then define a set of desirable axioms for additive feature attribution methods inspired by Shapley values:

\textbf{Local Accuracy} describes the ability of the explanation model to be accurate when evaluated at the target data point. For a model $f$ and its explanation model $f_x$ it should hold that: $f(x) = f_x(x')$.

\textbf{Missingness} describes that the contribution of a feature that is already missing in the simplified input $x'$, should be zero: $x_j = 0 \Rightarrow \phi_j = 0$.

\textbf{Consistency} means that if a feature has less impact (or stronger negative impact) on a second model, then on the original model, the contribution assigned to that feature for the second model should be smaller then that of the original model:
For a simplified input $z'$ let $z' / j$ be $z'$ modified s.t. $z'_j = 0$. For two models $f$ and $f'$ and their explanation models $f_x$ and $f'_x$ it should hold that:
\begin{align}
f'_x(z') - f'_x(z' / j) &\leq f_x(z') - f_x(z / j) \text{ for all } z'\in \{0,1\}^n \nonumber \\
\Rightarrow \phi_j(f', x) &\leq \phi_j(f, x)
\end{align}
Lundberg et al. \cite{lundberg2017unified} show that a unique function delivers the solution to these axioms:
\begin{align} \label{eq:SHAP}
    \phi_j(f,x) &:= \sum_{z'\subset x' \setminus j} G_{Mz'} \left(f_x(z' \cup j) - f_x(z')\right)\\
\text{with } G_{Mz'} &:=\frac{|z'|!(M-|z'|-1)!}{M!} \nonumber
\end{align}
The solution to this equation yields an additive feature attribution model that the authors call SHapley Additive exPlanations (SHAP). They then define different model specific and agnostic approaches to compute the SHAP values. 

We focus now on an approach designed for deep neural networks: DeepSHAP. 
Since the contributions $\phi_j(g,x)$ are easy to calculate for simple building blocks of neural networks, such as linear layers (e.g. convolutions, fully connected layers or normalization layers) and pointwise activation functions (e.g. ReLU, sigmoid), we just need to calculate the contribution of a full model, which is the concatenation of those layers, from the contributions of the individual layers.
For this sake the authors use an idea from DeepLIFT \cite{shrikumar2017learning}: Propagating Contributions through layers by a custom defined chain rule. For that they define multipliers as contributions relative to the input difference: 
\begin{equation}\label{eq:mult}
    m_{xf}^j := \frac{\phi_j(f,x)}{x_j - \E(x_j)}
\end{equation}
They then define a chain rule for these multipliers to propagate the contributions through the layers, similarly of how gradients are backpropagated from individual layer-gradients to calculate the gradient of the full model. Let $h=f\circ g$ with $g(y) = x \in \R^n$, then the chain rule states: 
\begin{align}
    m_{yh}^k = \sum_{j=1}^n m_{yg_j}^k m_{xf}^j
\end{align}
Applying this chain rule iteratively through all the layers makes it possible to compute the multipliers and then by inverting \autoref{eq:mult} also the contribution of the full neural network.

\section{Complex-valued Gradient based methods}\label{sec:cv_grad}
In this section, we describe how to adapt commonly used gradient based methods to the complex domain. Since the adaption is rather straight forward, we do not consider this a key contribution, but want to summarize the adaptations, since - to the best of our knowledge - this has not been done before. Additionally, these methods will provide a baseline for the experiments in \autoref{sec:eval} to compare Deep$\C$SHAP which is introduced in \autoref{sec:deepcshap}.
The adaption of all methods from \autoref{subsec:rv_grad} can be accomplished with the use of Wirtinger calculus. The Wirtinger derivative for a function $f: \C \to A$ with $A=\C \text{ or } \R$ are defined as:
\begin{align}
    \frac{\delta f}{\delta z} &:= \frac{1}{2} \left(\frac{\delta f}{\delta \real(z)} - i \frac{\delta f}{\delta \imag(z)}\right)\\
    \frac{\delta f}{\delta \overline{z}} &:= \frac{1}{2} \left(\frac{\delta f}{\delta \real(z)} + i \frac{\delta f}{\delta \imag(z)}\right)
\end{align}
The Wirtinger derivative w.r.t. the complex conjugate $\frac{\delta f}{\delta \overline{z}}$ has the property of pointing in the direction of the strongest ascend and the magnitude of the Wirtinger gradient is equal to the amount of change in that direction \cite{hirose2003complex}, just as the gradient does for real-valued functions.
Thus, replacing the real-valued gradient in the methods from \autoref{subsec:rv_grad} with $\frac{\delta f}{\delta \overline{z}}$ yields explanation methods with a similar interpretation as the methods in \autoref{subsec:rv_grad}. However, the Wirtinger gradient is complex-valued for complex-valued inputs and is thus not feasible to directly generate a saliency map.
The two straight forward options are to take either the magnitude of the complex gradient, or add up its real and imaginary part. The first variant has the drawback of neglecting the sign of the gradient, the second one may show a contribution of zero for non-zero gradients, if the real and imaginary part cancel out. We test both variants in \autoref{sec:eval}.

\begin{figure}[t]
    \centering
    \begin{subfigure}[c]{.23\linewidth}
    \begin{tikzpicture}[
    shortarrow/.style={
        shorten >= 1,
        shorten <= 1,
    },
    hatched/.style={
        pattern=north east lines,
        pattern color=lightgray,
    },
    filled/.style={
        circle,
        fill=black,
        draw,
        outer sep=0pt,
        inner sep=1.0pt
    },
    nonfilled/.style={
        circle,
        fill=white,
        draw,
        outer sep=0pt,
        inner sep=1.0pt
    },
    scale=2
]
    \draw [-, white] (0, -1) -- (0, 1) node {};
    \fill[hatched] (-1,1.1*\TickSize) rectangle (0,-1.1*\TickSize);

    \draw [->] (-1,0) -- (1,0) node {};
    \draw ($(0,0) + (0,\TickSize)$) -- ($(0,0) + (0,-\TickSize)$) node[below] {$0$};

    \draw (0, 0) node[nonfilled] (q0) {};

    \path[->] (-0.5, 0.05) edge [bend left, shortarrow, color=darkgray] node {} (q0);
    \path[->] (-0.5, -0.05) edge [bend right, shortarrow, color=darkgray] node {} (q0);

\end{tikzpicture}
    \caption{ReLU: Set unsuitable region ($x<0$) to zeros.}
    \end{subfigure}
    \hfill
    \begin{subfigure}[c]{.23\linewidth}
    \begin{tikzpicture}[
    shortarrow/.style={
        shorten >= 2,
        shorten <= 2,
    },
    hatched/.style={
        pattern=north east lines,
        pattern color=lightgray,
    },
    filled/.style={
        circle,
        fill=black,
        draw,
        outer sep=0pt,
        inner sep=1.0pt
    },
    nonfilled/.style={
        circle,
        fill=white,
        draw,
        outer sep=0pt,
        inner sep=1.0pt
    },
    scale=2
]
    \draw [-, white] (0, -1) -- (0, 1) node {};
    \fill[hatched] (-1,1) rectangle (0,-1);
    \fill[hatched] (0,0) rectangle (1,-1);

    \draw [->] (-1,0) -- (1,0) node {};
    \draw [->] (0,-1) -- (0,1) node {};

    \draw (0, 0) node[nonfilled] (q0) {};
    \path[->] (-0.5, 0.5) edge [shortarrow, color=darkgray] node {} (q0);
    \path[->] (-0.5, -0.5) edge [shortarrow, color=darkgray] node {} (q0);
    \path[->] (0.5, -0.5) edge [shortarrow, color=darkgray] node {} (q0);

\end{tikzpicture}
    \caption{zReLU: Set unsuitable region ($x \in \C_{2,3,4}$) to zeros.}
    \end{subfigure}
    \hfill
    \begin{subfigure}[c]{.23\linewidth}
    \begin{tikzpicture}[
    shortarrow/.style={
        shorten >= 1,
        shorten <= 1,
    },
    filled/.style={
        circle,
        fill=black,
        draw,
        outer sep=0pt,
        inner sep=1.0pt
    },
    nonfilled/.style={
        circle,
        fill=white,
        draw,
        outer sep=0pt,
        inner sep=1.0pt
    },
    scale=2
]
    \draw [-, white] (0, -1) -- (0, 1) node {};
    \draw [->] (-1,0) -- (1,0) node {};
    \draw ($(0,0) + (0,\TickSize)$) -- ($(0,0) + (0,-\TickSize)$) node[below] {$0$};

    \draw (-0.8, 0) node[filled] (p0) {};
    \draw (-0.5, 0) node[filled] (p1) {};
    \draw (0, 0) node[nonfilled] (q0) {};

    \path[->] (p0) edge [bend left, shortarrow] node {} (q0);
    \path[->] (p1) edge [bend left, shortarrow] node {} (q0);
\end{tikzpicture}
    \caption{ReLU: Projection $\R \to \R^+_0$.}
    \end{subfigure}
    \hfill
    \begin{subfigure}[c]{.23\linewidth}
    \begin{tikzpicture}[
    shortarrow/.style={
        shorten >= 2,
        shorten <= 2,
    },
    filled/.style={
        circle,
        fill=black,
        draw,
        outer sep=0pt,
        inner sep=1.0pt
    },
    nonfilled/.style={
        circle,
        fill=white,
        draw,
        outer sep=0pt,
        inner sep=1.0pt
    },
    scale=2
]
    \draw [-, white] (0, -1) -- (0, 1) node {};
    \draw [->] (-1,0) -- (1,0) node {};
    \draw [->] (0,-1) -- (0,1) node {};

    \draw (-0.8, 0.2) node[filled] (p0) {};
    \draw (-0.3, 0.5) node[filled] (p1) {};
    \draw (-0.5, -0.4) node[filled] (q0) {};
    \draw (-0.3, -0.7) node[filled] (q1) {};
    \draw (0.4, -0.5) node[filled] (r0) {};
    \draw (0.7, -0.2) node[filled] (r1) {};



    \foreach \i in {0,...,1}{
        \draw[->, shortarrow] (p\i) -- (p\i -| 0, 0) node {};
        \draw (p\i -| 0, 0) node[nonfilled] (s\i) {};
    }

    \foreach \i in {0,...,1}{
        \draw[->, shortarrow] (q\i) -- (0, 0) node {};
        \draw (0, 0) node[nonfilled] (t\i) {};
    }

    \foreach \i in {0,...,1}{
        \draw[->, shortarrow] (r\i) -- (r\i |- 0, 0) node {};
        \draw (r\i |- 0, 0) node[nonfilled] (u\i) {};
    }
\end{tikzpicture}
    \caption{$\C$ReLU: Projection $\C \to \C_1$}
    \end{subfigure}
    
    \caption{Two different interpretations of ReLU and their corresponding complex-valued adaptions. $\C_n$ denotes the $n$-th quadrant.}
    \label{fig:projection}
\end{figure}
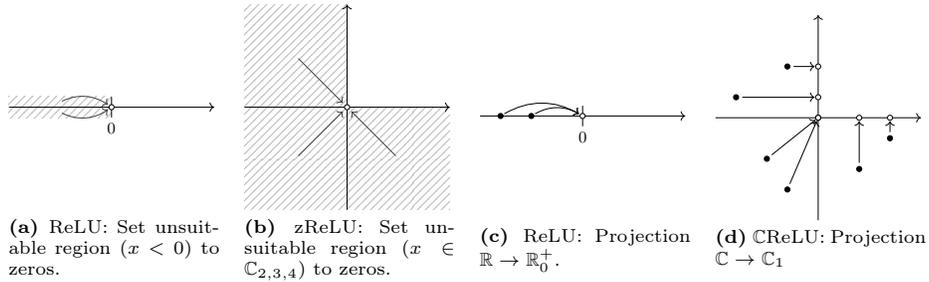

To adapt the Guided Backpropagation \cite{springenberg2014striving} it is necessary to define an equivalent formulation of "zeroing out negative gradients" in the backward pass for the Wirtinger gradients.
There are two different interpretations of the mapping: $ReLU(x) = max(0, x)$ in the real domain, which lead to different adaptions to the complex domain.
The first interpretation is to define a region of unsuitable values (in the real case all negative numbers) and interpret ReLU as the function setting all unsuitable (i.e. negative) values to zero.
Secondly, ReLU can be viewed as a projection: $\R$ to $\R^+_0$. Thus, every value in $\R$ is mapped to the closest value in $\R^+_0$, projecting negative values to zero and positive values to themselves.
These interpretations are visualized in \autoref{fig:projection} a) and c), respectively.

Following the first interpretation we propose to map every input to zero that is not in the first quadrant of $\C$ (i.e. real and imaginary parts are positive), thus defining the quadrants 2-4 as the region of unsuitable values (i.e. the equivalent to negative values in the real case).
Following the second of the two interpretations, we propose to project $\C$ to the first quadrant of $\C$, which yields the function $f(x) = max(0, \real(x)) + i \ max(0, \imag(x))$. Both variants are visualized in \autoref{fig:projection} b) and d), respectively.
The resulting formulations for complex-valued ReLU variants are in line with the ReLU variants for deep complex networks as proposed by Trabelsi et al. \cite{Trabels2018deep}, thus we call these variants Guided $z$-Backpropagation and Guided $\C$-Backpropagation, respectively.

\section{Deep$\C$SHAP}\label{sec:deepcshap}
This section marks the main contribution of this work. We aim to make DeepSHAP applicable to complex-valued neural networks since it is not applicable to complex-valued neural networks, due to its chain rule based on real-valued multipliers, as explained in \autoref{subsec:cr_dcshap}.
\subsection{$\C$SHAP values}
The first question that arises is how to define SHAP values in the complex-valued context, which is possible by simply adopting the definition from \autoref{eq:SHAP}: 
\begin{align}\label{eq:CV_SHAP}
    \phi_j(f,x) &:= \sum_{z'\subset x' \setminus j} G_{Mz'} \left(f_x(z' \cup j) - f_x(z')\right)\\
\text{with } G_{Mz'} &:=\frac{|z'|!(M-|z'|-1)!}{M!} \nonumber
\end{align}
Additionally, we define $\phi_0(f, x) := f_x(\emptyset)$ as the output of the model, if no feature is present.
Notably, we have that if $f: \C^n \to A$, with $A = \R \text{ or } \C$, then $\phi \in A$. Additionally, if $f: \C^n \to \C$ we can split the output into its real and imaginary parts and explain them separately, since it holds that:
\begin{align}
    \phi_j(\real(f), x) &= \real(\phi_j(f, x)) \nonumber \\ \label{eq:split_out}
    \phi_j(\imag(f), x) &= \imag(\phi_j(f, x))
\end{align}
\subsection{Chain rule for Deep$\C$SHAP}\label{subsec:cr_dcshap}
As pointed out before, DeepSHAP is not suitable to be used for complex-valued neural networks \cite{mayer2022deep}. The reason is that the multiplier defined in \autoref{eq:mult} is not suitable in complex-valued scenarios since complex-valued neural networks are usually not holomorphic (i.e. not complex differentiable) due to real output neurons or non-holomorphic activation functions. The quotient as defined in \autoref{eq:mult} does not converge for $(x_j - \E(x_j)) \to 0$ and is thus not suitable for small perturbations.

A similar problem arises in the optimization of complex-valued neural networks, which is solved by using the Wirtinger derivatives for optimization instead of the complex derivative.
The Wirtinger calculus motivates an alternative formulation of multipliers and a complex-valued chain rule that can be applied to complex-valued neural networks.

To define this complex-valued chain rule, we have to first define \textit{partial contributions} of the real and imaginary part. This definition is motivated by the following equation:
\begin{align}
    \phi_j(f,x) &= \sum_{z'\subset x' \setminus j} G_{Mz'}\left(f_x(z' \cup j) - f_x(z')\right) \\
    &= \sum_{z'\subset x' \setminus j} G_{Mz'} (f_x(z' \cup j) - f_x(z' \cup \real(j))  \nonumber \\
    &+ f_x(z' \cup \real(j)) - f_x(z'))\\
    &= \sum_{z'\subset x' \setminus j} G_{Mz'} \left(f_x(z' \cup j) - f_x(z' \cup \real(j))\right)\nonumber \\
    &+ \sum_{z'\subset x' \setminus j} G_{Mz'} \left( f_x(z' \cup \real(j)) - f_x(z')\right)\label{eq:cont_split}
\end{align}
where $\real(j)$ means to have only the real part of the $j$-th feature present, while the imaginary part stays absent. 
We now define the \textit{partial contributions} as the two summands of \autoref{eq:cont_split}:
\begin{align}\label{eq:SHAP_Rpart}
    \phi_{\real(j)}(f, x) &= \sum_{z'\subset x' \setminus j} G_{Mz'} \left( f_x(z' \cup \real(j)) - f_x(z')\right)\\\label{eq:SHAP_Ipart}
    \phi_{\imag(j)}(f, x) &= \sum_{z'\subset x' \setminus j} G_{Mz'} \left(f_x(z' \cup j) - f_x(z' \cup \real(j))\right)
\end{align}
Intuitively, the real partial contributions of a feature can be seen as the contributions of the real part of $x$, if the imaginary part of that feature is missing and the imaginary partial contribution as the contribution of the imaginary part, when the real part is never missing.
These quantities themselves do not hold meaningful information but are important for the formulation of the chain rule later on since we can use them to define the \textit{partial multipliers}:
\begin{align}\label{eq:RI_mults}
    m^{\real(j)}_{xf} = \frac{\phi_{\real(j)}(f, x)}{\real(x_j - \E(x_j))}; \hspace{2mm} m^{\imag(j)}_{xf} = \frac{\phi_{\imag(j)}(f, x)}{\imag(x_j - \E(x_j))}
\end{align}
Inspired by Wirtinger calculus, we can now define the actual multipliers used for the chain rule:
\begin{align}
    m^{j}_{xf} = \frac{1}{2} \left(m^{\real(j)}_{xf} - i \ m^{\imag(j)}_{xf} \right) \\
    m^{j}_{\overline{x}f} = \frac{1}{2} \left(m^{\real(j)}_{xf} + i \ m^{\imag(j)}_{xf} \right)
\end{align}
Before proceeding to the chain rule, we want to note two properties of the multipliers that will come in handy later on to reduce the computational load:
\begin{align}
    f: \C \to \R &\Rightarrow m^{j}_{\overline{x}f} = \overline{m^j_{xf}} \label{eq:prop_rv} \\
    f \text{ linear } &\Rightarrow m^j_{xf} = \frac{\delta f}{\delta x_j} \text{ and }  m^j_{\overline{x}f} = \frac{\delta f}{\delta \overline{x_j}} \label{eq:prop_linear}
\end{align}
where the derivative operators in the second property are meant as Wirtinger operators.
Proofs for these properties as well as for the now following chain rule can be found in the supplementary material. The complex-valued chain rule for a concatenation $h=f\circ g$ with $g(y) = x \in \C^n$ is:
\begin{align}\label{eq:cr_x}
m^{k}_{yh} &= \sum_{j=1}^n (m^k_{yg_j} m^j_{xf} + \overline{m^k_{yg_j}} m^j_{\overline{x}f}) \\ \label{eq:cr_olx}
m^{k}_{\overline{y}h} &= \sum_{j=1}^n (m^k_{\overline{y}g_j} m^j_{xf} + \overline{m^k_{\overline{y}g_j}} m^j_{\overline{x}f})
\end{align}
Since we are interested in the contributions rather than the multipliers, we have to reconstruct them after the use of the chain rule:
\begin{align}
    \phi_{\real(k)}(h, x) &= (m^{k}_{yh} + m^{k}_{\overline{y}h}) \real(x_k - \E(x_k)) \\
    \phi_{\imag(k)}(h, x) &= (m^{k}_{yh} - m^{k}_{\overline{y}h}) i\imag(x_k - \E(x_k)) \\
    \phi_k(h, x) &= \phi_{\real(k)}(h, x) + \phi_{\imag(k)}(h, x) 
\end{align}                  
We have summarized how to apply the chain rule to calculate the contributions of a concatenation of functions in Algorithm 1.

Note that by \autoref{eq:split_out} we can always assume the outer function $f$ to be real-valued since if it is not, we can just calculate the contributions to the real and imaginary part separately by replacing $f$ with $\real(f)$ and $\imag(f)$. Thus, we can make use of \autoref{eq:prop_rv} to simplify the calculation of the chain rule.

For numerical stability, we adopt an adaptation from DeepSHAP: If the denominator in the fraction in the multipliers in \autoref{eq:RI_mults} is close to zero (in our case smaller than $10^{-6}$), we replace it with the respective Wirtinger derivative.

Since the for-loop in Algorithm 1 can be vectorized and all operations can be defined as matrix multiplications, the algorithm can be efficiently computed on a GPU using deep learning frameworks (in our case: Pytorch). The contributions can then be computed in a single backward pass within milliseconds, see appendix for a runtime analysis.

\subsection{Max$\C$SHAP}\label{subsec:maxcshap}
To apply the chain rule from \autoref{subsec:cr_dcshap} we need to pre calculate the (partial) contributions $\phi_\real$ and $\phi_\imag$ for every layer. \cite{lundberg2017unified} derives SHAP values for real-valued max pooling layers, however their algorithm is not suitable for complex-valued max pooling. Since the complex numbers are not ordered, max pooling is commonly done by ordering by the magnitude of the patch, thus for an image patch $\mathcal{X} = \{x_1, \hdots, x_n\}$, we can define complex max pooling as
\begin{align}
    \C \text{maxpool} = \argmax_{x\in \mathcal{X}} |x|
\end{align}
In Algorithm 2 we define an algorithm to efficiently compute the contributions of a max pooling layer. For a presorted input (which can be accomplished in $O(n \log_2 n)$) the complexity of the algorithm is $O(n^2)$.


\begin{figure}[t]
\begin{subfigure}[t]{0.56\textwidth}
\textbf{Algorithm 1}

    \noindent
	Complex-valued chain rule for contributions $\phi_k(h,x)$ of $h=f\circ g$:
 \label{alg:cr_cont}
\begin{algorithmic}
	\State $\text{Analytic. calc }\phi_{\real(k)}(g, y), \phi_{\imag(k)}(g, y)$  \Comment{$*$}
	\State $ m^{\real(k)}_{yg} \gets \frac{\phi_{\real(k)}(g, y)}{\real(y_k - \E(y_k))} $
	\State $ m^{\imag(k)}_{yg} \gets \frac{\phi_{\imag(k)}(g, y)}{\imag(y_k - \E(y_k))} $
    \State $m^{k}_{yg} \gets \frac{1}{2} \left(m^{\real(k)}_{yg} - i \ m^{\imag(k)}_{yg} \right) $
    \State $m^{k}_{\overline{y}g} \gets \frac{1}{2} \left(m^{\real(k)}_{yg} + i \ m^{\imag(k)}_{yg} \right) $
    \For{$j=0$ \textbf{to} $n$}
    \State $\text{Analytic. calc } \phi_{\real(j)}(f, x), \phi_{\imag(j)}(f, x)$ \Comment{$*$}
	\State $ m^{\real(j)}_{xf} \gets \frac{\phi_{\real(j)}(f, x)}{\real(x_j - \E(x_j))} $
	\State $ m^{\imag(j)}_{xf} \gets \frac{\phi_{\imag(j)}(f, x)}{\imag(x_j - \E(x_j))} $
    \State $m^{j}_{xf} \gets \frac{1}{2} \left(m^{\real(j)}_{xf} - i \ m^{\imag(j)}_{xf} \right)$
    \State $m^{j}_{\overline{x}f} \gets \frac{1}{2} \left(m^{\real(j)}_{xf} + i \ m^{\imag(j)}_{xf} \right)$
    \EndFor
    \State $m^{k}_{yh} \gets \sum_{j=1}^n (m^k_{yg_j} m^j_{xf} + \overline{m^k_{yg_j}} m^j_{\overline{x}f})$
    \State $m^{k}_{\overline{y}h} \gets \sum_{j=1}^n (m^k_{\overline{y}g_j} m^j_{xf} + \overline{m^k_{\overline{y}g_j}} m^j_{\overline{x}f})$
    \State $\phi_{\real(k)}(h, x) \gets (m^{k}_{yh} + m^{k}_{\overline{y}h}) \real(x_k - \E(x_k))$
    \State $\phi_{\imag(k)}(h, x) \gets (m^{k}_{yh} - m^{k}_{\overline{y}h}) \imag(x_k - \E(x_k))$
    \State $\phi_k(h, x) \gets \phi_{\real(k)}(h, x) + \phi_{\imag(k)}(h, x)$\\
\end{algorithmic}
    $*$ easy and fast for building blocks of CVNNs (e.g. linear layers, 1-to-1 activation etc.)
\end{subfigure}
\vrule
\hfill
\begin{subfigure}[t]{0.40\textwidth}
\textbf{Algorithm 2}

\noindent
	Contributions $\phi_k$ of $\C$max pooling:\label{alg:maxcshap}
\begin{algorithmic}
\State Precomputed vector $M$ \Comment{*}
\State Input $X = (X_1, \hdots, X_n)$
\State Reference $Y = (Y_1, \hdots, Y_n)$
\State $m \gets \argmax_p(|X|,|Y|)$ \Comment{**}
\State $s \gets \argmax(\argmin_p(|X|, |Y|))$
\For{$k=0$ \textbf{to} $n-1$}
\State $m_x[k] \gets \argmin(X[k], s)$
\State $m_y[k] \gets \argmin(Y[k], s)$
\State $z_x \gets \sort(m_x, \text{key}=|\cdot|)$
\State $z_y \gets \sort(m_y, \text{key}=|\cdot|)$
\State $\phi_k \gets \langle z_x, M \rangle - \langle z_y, M \rangle$
\EndFor
\end{algorithmic}
    $*$ constant vector $M$ only depends on $n$ and can be pre computed for every max pooling layer. It incorporates the weighting factors $G_{Mz'}$ from \autoref{eq:CV_SHAP}.\\
    $**$ $\argmax_p$ denotes pointwise argmax
\end{subfigure}
\end{figure}

\section{Evaluation}\label{sec:eval}
We evaluate our method Deep$\C$SHAP in two experiments: An experiment on MNIST \cite{lecun1998mnist} as designed by Shrikumar et al. \cite{shrikumar2017learning} and already presented for the evaluation of the real-valued DeepSHAP \cite{lundberg2017unified}.
Additionally, we evaluate with a self designed experiment on a complex-valued PolSAR dataset \cite{asiyabi2023complex}. Details on the training procedure for both experiments can be found in the supplementary material.
Since no other explanation methods for complex-valued neural networks have been proposed, there is no competitive baseline, we could compare to. Comparing to an explanation algorithm for real-valued neural networks is not suitable either, since when comparing explanation methods between different models (e.g. a real-valued and a complex-valued to be able to apply the respective explanation method) we would not know, if a difference in our experiments has to be attributed to the difference in models, or the difference in explanation methods. Thus we treat the adaptations of gradient based methods in \autoref{sec:cv_grad} as baseline and compare our proposed Deep$\C$SHAP against those.

Additional to evaluating explanation quality, we also test our theoretical results from \autoref{sec:deepcshap} by verifying core properties of Deep$\C$SHAP computationally.
\begin{figure}[t]
    \centering
\includegraphics[width=\linewidth]{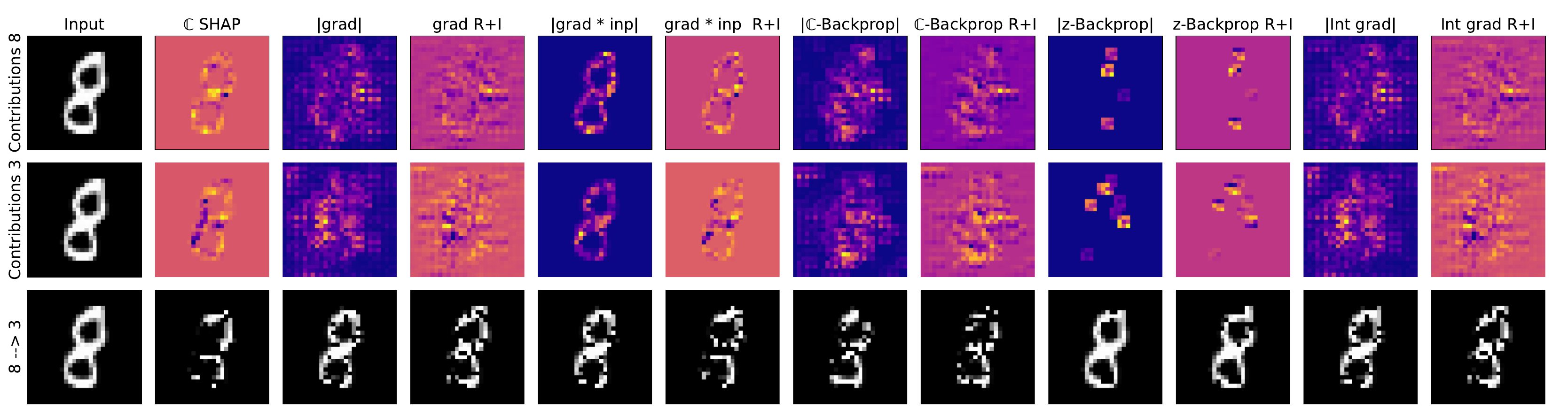}
    \caption{Qualitative results for MNIST experiment. Explanation scores for output 8 and 3 are shown and the image after masking 20\% of the most important pixels for both predictions. All images are normalized in itself, normalization over all methods is not suitable due to different orders of magnitude.}
    \label{fig:mnist_qual}
\end{figure}

\subsection{MNIST experiment}
\begin{figure}[t]
    \centering
\includegraphics[width=\linewidth]{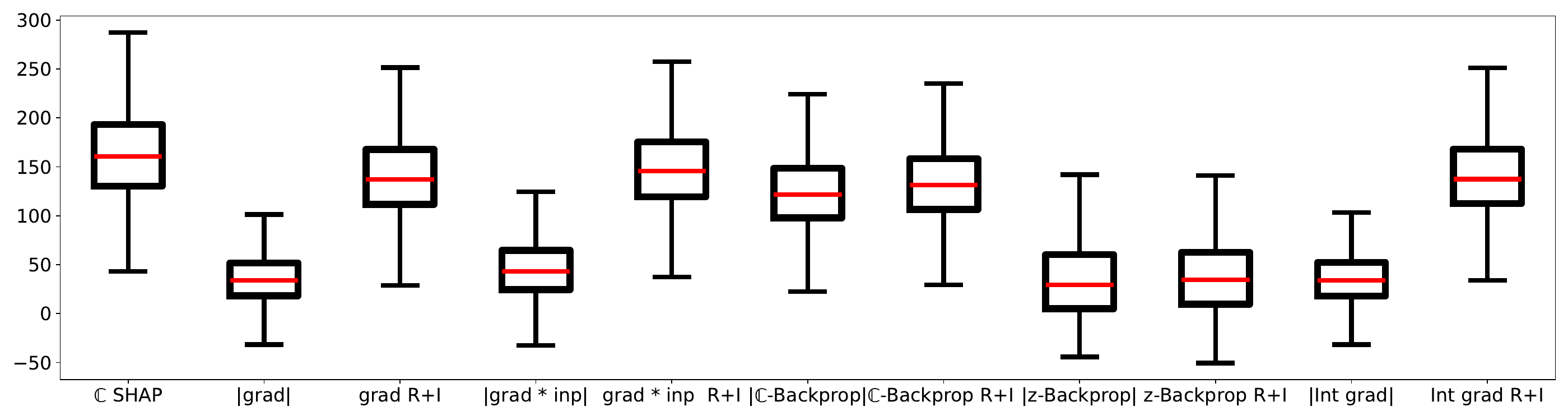}
    \caption{Quantitative results on MNIST experiment. Change in log-odds after masking 20\% of the most relevant features for classes 8 and 3 according to different explanation algorithms. The median is shown in red.}
    \label{fig:mnist_quan}
\end{figure}

We use a small complex-valued CNN with three blocks of $\C$ convolutions, $\C$Relu activations and $\C$ max pooling followed by a 2 layer MLP as a classifier on the MNIST dataset. The model obtains a 88\% accuracy on the digit classification task on the test set of MNIST. We adopt the experimental setup as performed by \cite{shrikumar2017learning, lundberg2017unified}: For a preset source class we randomly select an image from the test set and use the explanation methods to explain the source class output and the output for a fixed target class.
We then identify up to 20\% of the most important pixels for classifying the class correctly into the source class rather then into the target class by choosing the pixels with the highest explanation scores and remove these pixels by setting them to zero.
We then input the modified image into the model and evaluate the change in log-odds compared to the initial input.
A high score thus indicates the ability of the explanation algorithm to identify the most important features for a classification into a specific class.
Qualitative results are visualized in \autoref{fig:mnist_qual}. All explanations are very blurry except Deep$\C$SHAP and $|grad * input|$, which are the only methods whose explanations focus on the actual content of the image. This indicates that these methods assign contribution to the most important pixels. 
This is also visible in the quantitative results in \autoref{fig:mnist_quan}. There it shows that Deep$\C$SHAP performs best on this task. This result is in line with the results in \cite{lundberg2017unified}, where a similar result was shown for a real-valued neural network.
While none of the gradient based methods is on par with Deep$\C$SHAP, gradient times input and integrated gradients in the variant with the sum of real and imaginary parts show similarly good results. It is evident from these results, that the absolute value of the gradient leads to a worse performing explanation method, than the sum of real and imaginary part. The former performes very poorly for all methods except Guided$\C$Backpropagation. 
As for the two variants for Guided Backpropagation, the $\C$-ReLU version performs meaningfully better then the $z$-ReLU variant.


\subsection{PolSAR experiment}\label{subsec:ex_polsar}
\begin{figure}[t]
    \centering
\includegraphics[width=.8\linewidth]{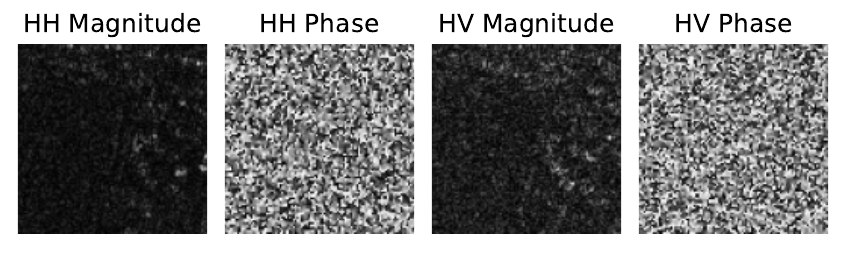}
    \caption{Example image from the PolSAR S1SLC$\_$CVDL dataset. Magnitude and Phase information of the complex-valued two channel (HH and HV) images are shown.}
    \label{fig:polsar_example}
\end{figure}

For this experiments we trained a complex-valued version of the ResNet18 \cite{he2016deep} with an Adam optimizer \cite{kingma2014adam} on the S1SLC$\_$CVDL \cite{asiyabi2023complex} datasets.
It consists of 276571 complex-valued PolSAR image patches of size $100\!\times\!100$ from three north American city areas.
They are labeled with seven classes: Agriculture, Forest, High-Density Urban, High-Rise Urban, Low-Density Urban, Industrial and Water. Per image patch two complex-valued channels, HH and HV polarization, are available. As can be seen in \autoref{fig:polsar_example}, these images are very noisy and thus a visual explanation is not feasible.

To still be able to evaluate explanation quality on this dataset, we designed an experiment on this task: We take two patches and input the four channels into the complex-valued neural network.
As output we expect the classes of both patches, making this a multiclass classification problem. The trained model obtains an accuracy of 93\% on this task (detailed results in the supp. material). 

If we explain the output log-odds for the one of the two correct labels, we know that it should be explained by the two corresponding input channels and the other two input channels should have no or a negative influence on the prediction.
We then measure to which extent this is fulfilled for a specific image by this measure:
Let $X^1, X^2, Y^1, Y^2$ be the four input channels from two patches from different classes. Let $\phi_k(f, Z)$ be the explanation by an explanation method of pixel $k \in \{1,...,n\}$ in channel $Z$. We then calculate:
\begin{equation}
    \frac{\displaystyle\sum_{l=1}^2\sum_{k=0}^n \phi_k(f,X^l)}{\displaystyle\sum_{l=1}^2\sum_{k=0}^n \phi_k(f,X^l) + \sum_{l=1}^2\sum_{k=0}^n \phi_k(f,Y^l)}
\end{equation}
For strictly positive explanations, this measure is bounded at 1 and would signal that all important features are in the correct channels. For possibly negative values of signed explanation method (i.e. Deep$\C$SHAP and the signed gradient variants) a value greater than 1 indicates that the incorrect channels contributed negatively to the final decision, which is desirable (since these channels show an image of a different class). Hence a score $\geq 1$ is desirable.

\begin{figure}[t]
    \centering
\includegraphics[width=\linewidth]{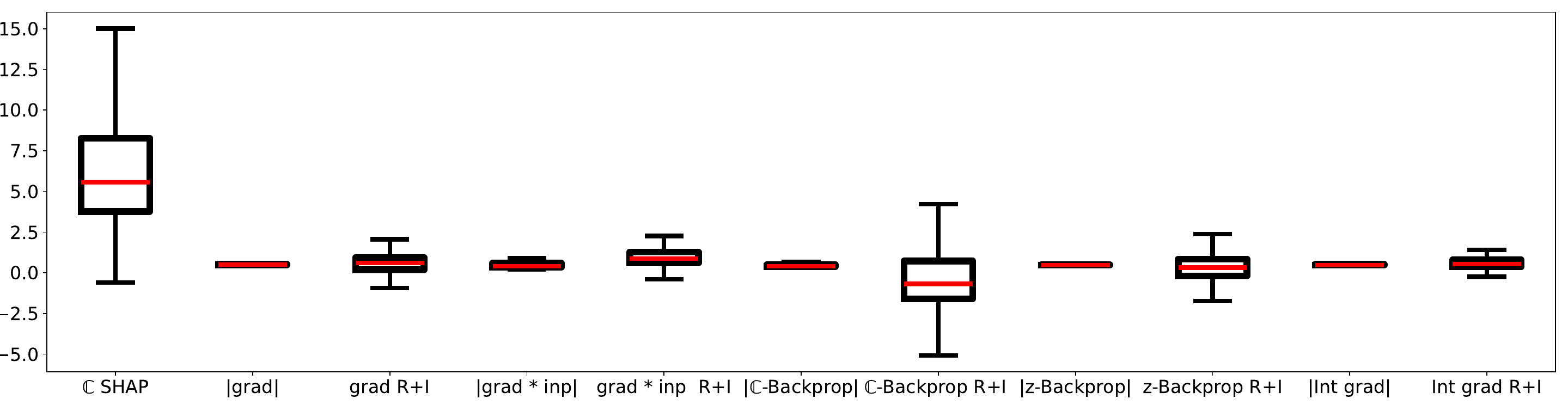}
    \caption{Quantitative results for the PolSAR experiment. Fraction of calculated contributions from the correct channels versus all channels are visualized for all explanation methods introduced in this paper. For gradient based methods $|\cdot|$ denotes the absolute value of the respecting method, R+I denotes adding up real and imaginary part.}
    \label{fig:polsar_quan}
\end{figure}

\autoref{fig:polsar_quan} shows that Deep$\C$SHAP performs best on this task. All the gradient based methods perform relatively poorly. The results from the MNIST experiment are mostly confirmed, as the $\mathcal{R}+\mathcal{I}$ variants of gradient times input and integrated gradients perform best. Overall, the absolute value  with median scores around 0.5, indicating that they marked the features of the incorrect channels to be equally important as the features of the correct channels. 
All variants of the Guided Backpropagation perform poorly, while the simple gradient explanation, gradient multiplied by input and integrated gradient show median scores around 1 when the real and imaginary parts are added. 

\subsection{Testing SHAP axioms}

To verify our theoretical results from \autoref{sec:deepcshap} we test Missingness and Local Accuracy as two of the axioms of the SHAP framework.

To test Missingness, we report the percentage of missing inputs that are assigned a value other than $0$ by Deep$\C$SHAP. To test Local Accuracy, we report the average of the following equation over all outputs and the whole test set. Let $X$ be an input image and $f$ the prediction model:
\begin{equation}\label{eq:loc_acc}
    \left|\left(\sum_{j=1}^n \phi_j(f, X) + \phi_0\right) - f(X)\right|
\end{equation}
As \autoref{tab:axioms} shows, Missingness is always fulfilled. The (absolute) Local Accuracy Error of $9.8 \times 10^{-6}$ and $3.5 \times 10^{-3}$ (resulting in a relative error of $2.3 \times 10^{-7}$ and $2.7 \times 10^{-5}$) on MNIST and S1SLC$\_$CVDL respectively can be explained by the adaptations made for numeric stability.

\begin{table}[t]
    \centering
\begin{tabular}{c|c|c}
     & \ \ MNIST \ \ & \ \ S1SLC$\_$CVDL \ \ \\
     \hline
\ Local Accuracy Error \ \  & $9.8 \times 10^{-6}$ & $3.5 \times 10^{-3}$ \\
    \hline
Average $|$output$|$   & 39.21 & 129.27 \\
    \hline
Missingness Error & 0.0\% & 0.0\% \\
    \hline 
    \multicolumn{1}{c}{\vspace{1mm}}
\end{tabular}
    \caption{Validation of Local Accuracy and Missingness in Deep$\C$SHAP. Local Accuracy error denotes the value from \autoref{eq:loc_acc}. Average absolute value of the model output is reported to make the results for Local Accuracy comparable. Missingness error denotes the fraction of missing features that are assigned a non-zero contribution.}.
    \label{tab:axioms}
\end{table}

\section{Conclusion}\label{sec:con}
In this paper we have introduced multiple explanation algorithms for complex-valued neural networks by adapting four gradient based explanation methods to the complex domain and by developing a complex-valued version of the well known DeepSHAP \cite{lundberg2017unified} explanation method, which we call Deep$\C$SHAP.
The major theoretical result to obtain the latter is the complex-valued chain rule for contributions as introduced in \autoref{sec:deepcshap}.
By providing these methods, we fill the gap of missing explanation methods for complex-valued neural networks, which are rising in popularity for applications with real- and complex-valued data types.

We analyze explanation quality on the real-valued MNIST dataset and the real world complex-valued PolSAR S1SLC$\_$CVDL dataset. We show that our method Deep$\C$SHAP outperforms the gradient based methods on both tasks.

Additionally, we have validated our theoretical results on these datasets. We show that the SHAP properties Local Accuracy and Missingness are fulfilled. 
This and the theoretical guarantees given by the theoretical foundation of Shapley values make this method suited to explain complex-valued neural networks.

As future work, we aim to adopt more explanation methods to the complex domain, such as LIME and kernelSHAP.
Additionally, applying Deep$\C$SHAP on different modalities will be interesting, especially modalities that are easily visually interpretable, such as MRI images to make use of the generated head maps, rather than quantitative results on the hard-to-understand PolSAR images. 


\bibliographystyle{splncs04}
\bibliography{bib}

\begin{thebibliography}{10}
\providecommand{\url}[1]{\texttt{#1}}
\providecommand{\urlprefix}{URL }
\providecommand{\doi}[1]{https://doi.org/#1}

\bibitem{adadi2018peeking}
Adadi, A., Berrada, M.: Peeking inside the black-box: a survey on explainable
  artificial intelligence ({XAI}). IEEE Access  \textbf{6},  52138--52160
  (2018)

\bibitem{asiyabi2023complex}
Asiyabi, R.M., Datcu, M., Anghel, A., Nies, H.: Complex-valued end-to-end deep
  network with coherency preservation for complex-valued {SAR} data
  reconstruction and classification. IEEE Transactions on Geoscience and Remote
  Sensing  (2023)

\bibitem{bach2015pixel}
Bach, S., Binder, A., Montavon, G., Klauschen, F., M{\"u}ller, K.R., Samek, W.:
  On pixel-wise explanations for non-linear classifier decisions by layer-wise
  relevance propagation. PloS one  \textbf{10}(7),  e0130140 (2015)

\bibitem{barrachina2023impact}
Barrachina, J.A., Ren, C., Morisseau, C., Vieillard, G., Ovarlez, J.P.: Impact
  of {PolSAR} pre-processing and balancing methods on complex-valued neural
  networks segmentation tasks. IEEE Open Journal of Signal Processing
  \textbf{4},  157--166 (2023)

\bibitem{bassey2021survey}
Bassey, J., Qian, L., Li, X.: A survey of complex-valued neural networks.
  arXiv:2101.12249  (2021)

\bibitem{chen2023spectral}
Chen, H., He, F., Lei, S., Tao, D.: Spectral complexity-scaled generalisation
  bound of complex-valued neural networks. Artificial Intelligence
  \textbf{322},  103951 (2023)

\bibitem{choi2018phase}
Choi, H.S., Kim, J.H., Huh, J., Kim, A., Ha, J.W., Lee, K.: Phase-aware speech
  enhancement with deep complex {U-Net}. In: ICLR (2018)

\bibitem{cole2021analysis}
Cole, E., Cheng, J., Pauly, J., Vasanawala, S.: Analysis of deep complex-valued
  convolutional neural networks for {MRI} reconstruction and phase-focused
  applications. Magnetic Resonance in Medicine  \textbf{86}(2),  1093--1109
  (2021)

\bibitem{eilers2023building}
Eilers, F., Jiang, X.: Building blocks for a complex-valued transformer
  architecture. In: IEEE International Conference on Acoustics, Speech and
  Signal Processing (ICASSP) (2023)

\bibitem{he2016deep}
He, K., Zhang, X., Ren, S., Sun, J.: Deep residual learning for image
  recognition. In: IEEE Conference on Computer Vision and Pattern Recognition.
  pp. 770--778 (2016)

\bibitem{hirose2003complex}
Hirose, A.: Complex-Valued Neural Networks: Theories and Applications. World
  Scientific (2003)

\bibitem{hu2020dccrn}
Hu, Y., Liu, Y., Lv, S., Xing, M., Zhang, S., Fu, Y., Wu, J., Zhang, B., Xie,
  L.: {DCCRN}: Deep complex convolution recurrent network for phase-aware
  speech enhancement. In: INTERSPEECH (2020)

\bibitem{kindermans2016investigating}
Kindermans, P.J., Sch{\"u}tt, K., M{\"u}ller, K.R., D{\"a}hne, S.:
  Investigating the influence of noise and distractors on the interpretation of
  neural networks. arXiv:1611.07270  (2016)

\bibitem{kingma2014adam}
Kingma, D.P., Ba, J.: Adam: {A} method for stochastic optimization. In: 3rd
  International Conference on Learning Representations {ICLR} (2015)

\bibitem{ko2022coshnet}
Ko, M., Panchal, U.K., Andrade-Loarca, H., Mendez-Vazquez, A.: Coshnet: A
  hybird complex valued neural network using shearlets. arXiv:2208.06882
  (2022)

\bibitem{lecun1998mnist}
LeCun, Y.: The {MNIST} database of handwritten digits. http://yann. lecun.
  com/exdb/mnist/  (1998)

\bibitem{lee2022complex}
Lee, C., Hasegawa, H., Gao, S.: Complex-valued neural networks: A comprehensive
  survey. IEEE/CAA Journal of Automatica Sinica  \textbf{9}(8),  1406--1426
  (2022)

\bibitem{li2020sscv}
Li, X., Sun, Q., Li, L., Liu, X., Liu, H., Jiao, L., Liu, F.: {SSCV-GANs}:
  Semi-supervised complex-valued {GANs} for {PolSAR} image classification. IEEE
  Access  \textbf{8},  146560--146576 (2020)

\bibitem{lin2022sspnet}
Lin, Q.H., Niu, Y.W., Sui, J., Zhao, W.D., Zhuo, C., Calhoun, V.D.: {SSPNet}:
  An interpretable {3D-CNN} for classification of schizophrenia using phase
  maps of resting-state complex-valued {fMRI} data. Medical Image Analysis
  \textbf{79},  102430 (2022)

\bibitem{lundberg2017unified}
Lundberg, S.M., Lee, S.I.: A unified approach to interpreting model
  predictions. Advances in Neural Information Processing Systems  \textbf{30}
  (2017)

\bibitem{mayer2022deep}
Mayer, K.S., M{\"u}ller, C., Soares, J.A., de~Castro, F.C.C., Arantes, D.S.:
  Deep phase-transmittance {RBF} neural network for beamforming with multiple
  users. IEEE Wireless Communications Letters  \textbf{11}(7),  1498--1502
  (2022)

\bibitem{ribeiro2016should}
Ribeiro, M.T., Singh, S., Guestrin, C.: "{Why} should {I} trust you?"
  explaining the predictions of any classifier. In: 22nd ACM SIGKDD
  International Conference on Knowledge Discovery and Data Mining. pp.
  1135--1144 (2016)

\bibitem{shrikumar2017learning}
Shrikumar, A., Greenside, P., Kundaje, A.: Learning important features through
  propagating activation differences. In: International Conference on Machine
  Learning. pp. 3145--3153. PMLR (2017)

\bibitem{simonyan2013deep}
Simonyan, K., Vedaldi, A., Zisserman, A.: Deep inside convolutional networks:
  Visualising image classification models and saliency maps. In: 2nd
  International Conference on Learning Representations {(ICLR)} (2014)

\bibitem{springenberg2014striving}
Springenberg, J.T., Dosovitskiy, A., Brox, T., Riedmiller, M.A.: Striving for
  simplicity: The all convolutional net. In: 3rd International Conference on
  Learning Representations {(ICLR)} (2015)

\bibitem{vstrumbelj2014explaining}
{\v{S}}trumbelj, E., Kononenko, I.: Explaining prediction models and individual
  predictions with feature contributions. Knowledge and Information Systems
  \textbf{41},  647--665 (2014)

\bibitem{sundararajan2016gradients}
Sundararajan, M., Taly, A., Yan, Q.: Gradients of counterfactuals.
  arXiv:1611.02639  (2016)

\bibitem{tan2022real}
Tan, Z.H., Xie, Y., Jiang, Y., Zhou, Z.H.: Real-valued backpropagation is
  unsuitable for complex-valued neural networks. Advances in Neural Information
  Processing Systems  \textbf{35},  34052--34063 (2022)

\bibitem{Trabels2018deep}
Trabelsi, C., Bilaniuk, O., Zhang, Y., Serdyuk, D., Subramanian, S., Santos,
  J.F., Mehri, S., Rostamzadeh, N., Bengio, Y., Pal, C.J.: Deep complex
  networks. In: 6th International Conference on Learning Representations
  {(ICLR)} (2018)

\bibitem{vasudeva2022compressed}
Vasudeva, B., Deora, P., Bhattacharya, S., Pradhan, P.M.: Compressed sensing
  {MRI} reconstruction with {Co-VeGAN}: Complex-valued generative adversarial
  network. In: WACV. pp. 1779--1788 (2022)

\bibitem{yang2020complex}
Yang, M., Ma, M.Q., Li, D., Tsai, Y.H.H., Salakhutdinov, R.: Complex
  transformer: A framework for modeling complex-valued sequence. In: IEEE
  International Conference on Acoustics, Speech and Signal Processing (ICASSP).
  pp. 4232--4236 (2020)

\bibitem{yin2020phasen}
Yin, D., Luo, C., Xiong, Z., Zeng, W.: Phasen: A phase-and-harmonics-aware
  speech enhancement network. In: AAAI Conference on Artificial Intelligence.
  vol.~34, pp. 9458--9465 (2020)

\bibitem{zhang2021optical}
Zhang, H., Gu, M., Jiang, X., Thompson, J., Cai, H., Paesani, S., Santagati,
  R., Laing, A., Zhang, Y., Yung, M., et~al.: An optical neural chip for
  implementing complex-valued neural network. Nature Communications
  \textbf{12}(1), ~457 (2021)

\end{thebibliography}

\clearpage
\setcounter{page}{1}
\maketitlesupplementary


\section{Proof of properties of complex-valued multipliers}
We want to prove the following properties for complex-valued multipliers as mentioned in the main paper in Equations \ref{eq:prop_rv} - \ref{eq:prop_linear}:
\begin{align}
    f: \C \to \R &\Rightarrow m^{j}_{\overline{x}f} = \overline{m^j_{xf}} \label{eq:prop_rv_sup} \\
    f \text{ linear } &\Rightarrow m^j_{xf} = \frac{\delta f}{\delta x_j} \text{ and }  m^j_{\overline{x}f} = \frac{\delta f}{\delta \overline{x_j}} \label{eq:prop_linear_sup}
\end{align}
    \autoref{eq:prop_rv_sup} follows from the fact that if $f:\C \to \R$, then $\phi_j(\real(f), x)$ and $\phi_j(\imag(f), x)$ are real-valued and then also $m^{\real(j)}_{xf}$ and $m^{\imag(j)}_{xf}$. Thus it holds that:
    \begin{align}
    m^{j}_{\overline{x}f} &= \frac{1}{2}\left(m^{\real(j)}_{xf} + i \ m^{\imag(j)}_{xf}\right) \\
     &= \overline{ \frac{1}{2}\left(m^{\real(j)}_{xf} - i \ m^{\imag(j)}_{xf}\right)} \\
     &= \overline{m^j_{xf}}
    \end{align}
    To prove \autoref{eq:prop_linear_sup} let $f(x) := Ax$ with $A \in \C^{1\times n}$ (for multidimensional output, we analyze each output separately and thus every row of $A$ separately), then with \autoref{eq:SHAP_Rpart} / \ref{eq:SHAP_Ipart} and $\Delta x := x - \E(x)$, we get:
    \begin{align}
       \phi_{\real(j)}(f,x) &= A_j \real(\Delta x_j) \text{ and }\\
       \phi_{\imag(j)}(f,x) &= i A_j  \imag(\Delta x_j)
    \end{align}
    and thus:
    \begin{align}
       m^{\real(j)}_{xf} &= \frac{A_j \real(\Delta x_j)}{\real(\Delta x_j)} = A_j \text{ and }\\
       m^{\imag(j)}_{xf} &= \frac{i A_j \imag(\Delta x_j)}{\imag(\Delta x_j)} = i A_j
    \end{align}
    It then follows:
    \begin{align}
        m^{j}_{xf} &=\frac{1}{2} (A_j - i^2 A_j) = A_j = \frac{\delta f}{\delta x_j}\\
        m^{j}_{\overline{x}f} &=\frac{1}{2} (A_j + i^2 A_j) = 0 = \frac{\delta f}{\delta \overline{x_j}}
    \end{align}
    
    \hfill $\Box$

\section{Proof of chain rule}\label{asec:proof}
To prove the chain rule for complex-valued multipliers (Equations \ref{eq:cr_x} and \ref{eq:cr_olx} in the main paper), we have to show:
\begin{align}
    \sum_{k\leq m}  \phi_k(h,y) = h(y) - \phi_0
\end{align}
when applying the chain rule for complex-valued multipliers. For an easier notation let $\Delta z = z - \E(z)$. Then it holds:
\begin{align}
    & \sum_{k\leq m}  \phi_k(h,y)  \\
    = & \sum_{k\leq m} (\phi_{\real(k)}(h, y) + \phi_{\imag(k)}(h, y))\\
    = & \sum_{k\leq m} m^{\real(k)}_{yh} \real(\Delta y_k) + m^{\imag(k)}_{yh} \imag(\Delta y_k)\\
    = & \sum_{k\leq m} m^{\real(k)}_{yh} \real(\Delta y_k) + m^{\imag(k)}_{yh} \imag(\Delta y_k)\\
    = & \sum_{k\leq m} (m^k_{\overline{y}h} + m^k_{yh}) \real(\Delta y_k) + (m^{k}_{yh} - m^k_{\overline{y}h}) i \ \imag(\Delta y_k)\\
    = & \sum_{k\leq m} \sum_{j\leq n} ((m^k_{\overline{y}g_j} m^j_{xf} + \overline{m^k_{\overline{y}g_j}} m^j_{\overline{x}f}) \\
    & + (m^k_{yg_j} m^j_{xf} + \overline{m^k_{yg_j}} m^j_{\overline{x}f})) \real(\Delta y_k)\\
    & + ((m^k_{yg_j} m^j_{xf} + \overline{m^k_{yg_j}} m^j_{\overline{x}f}) \\
    & - (m^k_{\overline{y}g_j} m^j_{xf} + \overline{m^k_{\overline{y}g_j}} m^j_{\overline{x}f}) )i\ \imag(\Delta y_k)\\
    = & \sum_{k\leq m} \sum_{j\leq n} ((m^{\real(k)}_{yg_j} m^{j}_{xf}  + \overline{m}^{\real(k)}_{yg_j} m^{j}_{\overline{x}f})  \real(\Delta y_k)\\
    & + (m^{\imag(k)}_{yg_j} m^{j}_{xf}  + \overline{m}^{\imag(k)}_{yg_j} m^{j}_{\overline{x}f}) (-i)i\ \imag(\Delta y_k))\\
    = & \sum_{j\leq n} (m^j_{xf} (\Delta x_j) +  m^j_{\overline{x}f} \overline{(\Delta x_j)}) \\
    = & \sum_{j\leq n} ((m^j_{xf} + m^j_{\overline{x}f}) \real(\Delta x_j) +  (m^j_{xf} - m^j_{\overline{x}f}) \imag(\Delta x_j)) \\
    = & \sum_{j\leq n} \phi_{\real(j)}(f, x) + \phi_{\imag(j)}(f, x) \\
    = & \sum_{j\leq n} \phi_{j}(f, x) \\
    = & f(x) - \phi_0\\
    = & h(y) - \phi_0
    \end{align}
    \hfill $\Box$

\begin{table*}[ht!]
    \centering
\begin{tabular}{c|c|c|c|c|c|c|c}
     &  Agriculture & Forest & High Density & Buildings & Low Density & Industrial & Water\\
     & & & Urban & & Urban & & \\
     \hline
Precision  & 0.999 & 0.999 & 0.999 & 0.997 & 0.999 & 0.999 & 0.999\\
    \hline
Recall  & 0.941 & 0.946 & 0.680 & 0.722 & 0.650 & 0.863 & 0.999\\
    \hline
    \multicolumn{1}{c}{\vspace{1mm}}
\end{tabular}
    \caption{Per class precision and recall of the complex-valued ResNet18 trained on the S1SLC$\_$CVDL dataset.}.
    \label{tab:prec_rec}
\end{table*}

\begin{figure}[t]
    \centering
\includegraphics[width=\linewidth]{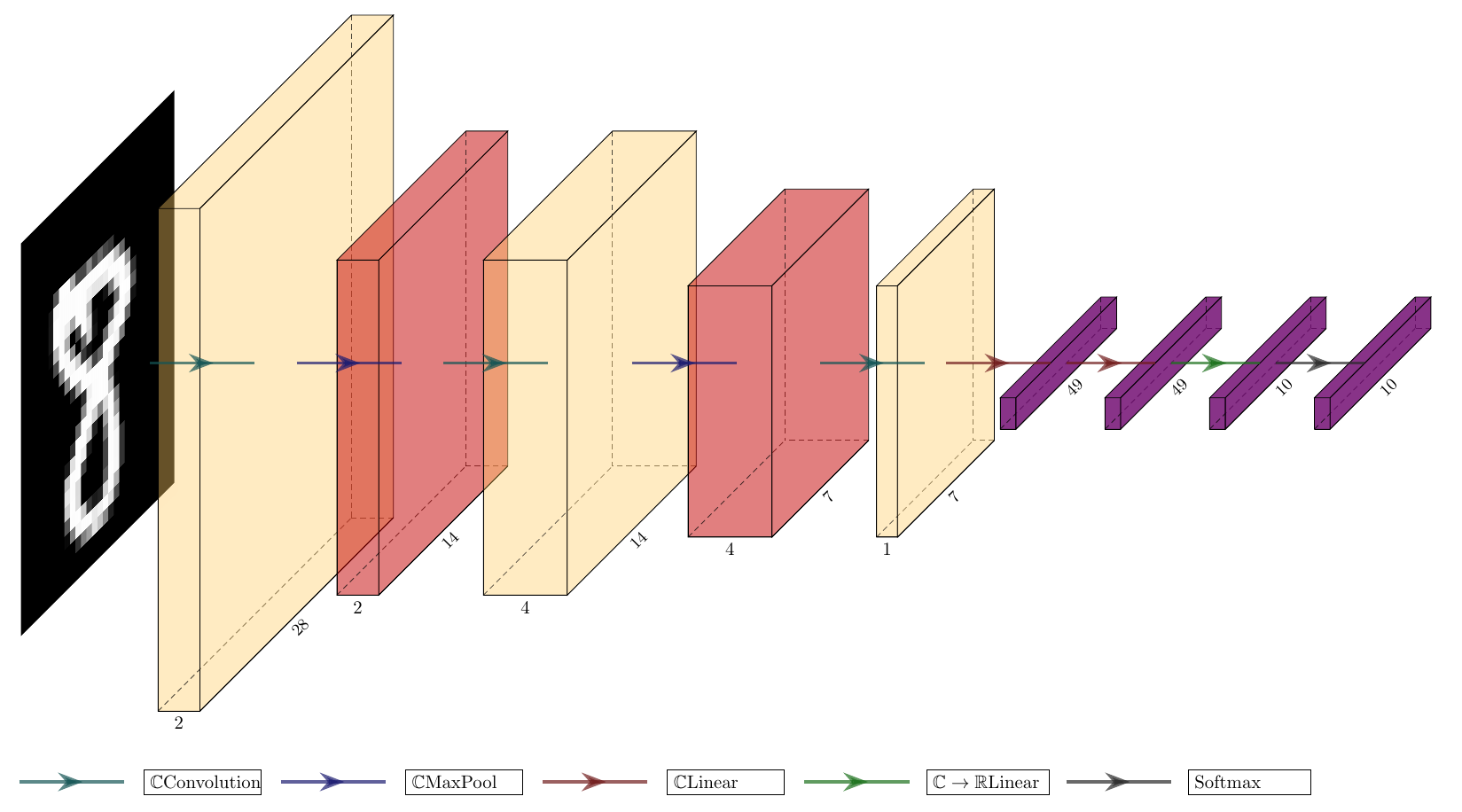}
    \caption{$\mathbb{C}$Convolutional model designed for the MNIST dataset task.}
    \label{fig:mnist_model}
\end{figure}

\section{Details on training and results -- PolSAR}
The experimental setup is explained in detail \autoref{subsec:ex_polsar}.
We used a complex-valued version of the ResNet18 \cite{he2016deep} with four input channels and 7 outputs classes, where the ground truth label has one or two classes, making this a multiclass classification problem. We trained the model with an Adam \cite{kingma2014adam} optimizer in pytorch. The model adaptation to the complex domain was made, as suggested by Trabelsi et al. \cite{Trabels2018deep}. The training was performed with the following hyperparameters:
\begin{itemize}
    \item Learning rate: 0.001
    \item Epochs: 20
    \item Loss: binary cross entropy
    \item Adam betas: 0.9, 0.999
\end{itemize}

The test accuracy of the trained model is 0.93, precision and recall for the individual classes can be seen in \autoref{tab:prec_rec}.
    
\section{Details on training and results -- MNIST}
We use a small complex-valued CNN with three blocks of $\C$ convolutions, $\C$Relu activations and $\C$ max pooling followed by a 2 layer MLP as a classifier on the MNIST dataset as can be seen in \autoref{fig:mnist_model}.
We train with an Adam \cite{kingma2014adam} optimizer in pytorch. The model was build from building blocks as suggested by Trabelsi et al. \cite{Trabels2018deep} with max pooling layers as described in \autoref{subsec:maxcshap}. The training was performed with the following hyperparameters:
\begin{itemize}
    \item Learning rate: 0.001
    \item Epochs: 20
    \item Loss: cross entropy
    \item Adam betas: 0.9, 0.999
\end{itemize}

The test accuracy of the trained model is 0.88.

\section{Runtime analysis of explanation methods}

\begin{table*}[ht!]
    \centering
\begin{tabular}{c|c|c|c|c|c|c}
     &  Deep$\C$SHAP & Grad & Grad*Input & $\C$-Backprop & z-Backprop & int. Grad\\
     \hline
MNIST & 132 & 5.0/4.7 & 4.7/4.7 & 5.3/5.3 & 5.2/5.4 & 22.6/22.6\\
    \hline
 S1SLC$\_$CVDL & 380 & 57.2/57.8 & 59.4/59.7 & 64.1/63.8 & 63.3/63.3 & 302/302\\
    \hline
    \multicolumn{1}{c}{\vspace{1mm}}
\end{tabular}
    \caption{Runtime of different explanation methods on both datasets in milliseconds. For gradient based methods (all but Deep$\C$SHAP) the first number shows the result for the version when using the absolute value, the second number when using $\real + \imag$.}
    \label{tab:runtime}
\end{table*}

For both experiments from \autoref{sec:eval} we performed a runtime analysis for all explaination methods presented in \autoref{sec:deepcshap} and \autoref{sec:cv_grad}. Results can be seen in \autoref{tab:runtime}. As expected, all the gradient based methods are very fast. Just the integrated gradient method is significantly slower than the other methods, which is expected, since for the the integration the gradient has to be calculated for multiple inputs. We use 5 integrands which corresponds with the slow down factor of $\sim 5$. As can be observed, Deep$\C$SHAP is the slowest of all the explanation methods. It is however still relatively fast, being able to explain $\sim 8$ images per second on MNIST and $\sim 3$ images per second on the S1SLC$\_$CVDL PolSAR dataset.

\end{document}